# Using Natural Language Processing to find Indication for Burnout with Text Classification: From Online Data to Real-World Data

Mascha Kurpicz-Briki[1], Ghofrane Merhbene[1], Alexandre Puttick[1], Souhir Ben Souissi[1], Jannic Bieri[2], Thomas Jörg Müller[3,4], Christoph Golz[2]

[1] Applied Machine Intelligence, Bern University of Applied Sciences, Biel, Switzerland
[2] School of Health Professions, Bern University of Applied Sciences, Bern, Switzerland
[3] Private Clinic Meiringen, Bern, Switzerland
[4] Translational Research Center, University Hospital of Psychiatry and Psychotherapy, University of Bern, Bern, Switzerland

## Abstract

**Background**: Burnout, classified as a syndrome in the ICD-11, arises from chronic workplace stress that has not been effectively managed. It is characterized by exhaustion, cynicism, and reduced professional efficacy, and estimates of its prevalence vary significantly due to inconsistent measurement methods. Recent advancements in Natural Language Processing (NLP) and machine learning offer promising tools for detecting burnout through textual data analysis, with studies demonstrating high predictive accuracy.

**Objective**: This paper contributes to burnout detection in German texts by: (a) collecting an anonymous real-world dataset including free-text answers and Oldenburg Burnout Inventory (OLBI) responses; (b) demonstrating the limitations of a GermanBERT-based classifier trained on online data; (c) presenting two versions of a curated BurnoutExpressions dataset, which yielded models that perform well in real-world applications; and (d) providing qualitative insights from an interdisciplinary focus group on the interpretability of AI models used for burnout detection.

**Methods**: We used data triangulation across three sources: (a) pre-existing online datasets, (b) manually curated synthetic datasets, and (c) a real-world dataset featuring free-text responses collected alongside results from OLBI, a research-validated burnout inventory. A pre-trained GermanBERT model was used, with a vocabulary that was augmented to include terms from the BurnoutExpressions dataset prior to fine-tuning for binary classification (burnout indication vs. no indication). Four classifiers were developed, each trained on different datasets with an 80/20 training and validation split. The models were subsequently evaluated using real-world data, consisting of free-text responses labeled with OLBI scores. Additionally, an explainability analysis was conducted, leveraging word attribution visualizations to interpret model decisions, including qualitative results from discussions with domain experts.

**Results:**
The classifier trained solely on online data showed poor performance in real-world testing despite high training scores. Classifiers fine-tuned on the combined dataset of online and synthetic data achieved the best F1-scores, indicating improved generalization. Models trained on the manually curated dataset also performed well, with marginal performance differences between the use of only curated expressions as training data and augmenting training data with AI-generated extensions.

**Conclusions**:
Our findings emphasize the need for greater collaboration between AI researchers and clinical experts to refine burnout detection models. Additionally, more real-world data is essential to validate and enhance the effectiveness of current AI methods developed in NLP research, which are often based on data automatically scraped from online sources and not evaluated in a real-world context. This is essential for ensuring AI tools are well suited for practical applications.

# Introduction

Stress is a complex concept with physiological and psychological components. It can be triggered by numerous stressors building up and creating an imbalance between environmental demands and one's ability to cope [1]. Stress refers to the interaction between a person and their environment, characterized by experiencing the environment as burdensome or overwhelming and perceiving a threat to their well-being [2]. This state of distress must be taken seriously, as it can significantly impair one's daily life [3]. Work-related stress is caused by job-related stressors and can lead to physical, behavioral, or psychological consequences, affecting both the employee's health and well-being, as well as that of the organization for which they work [4]. While limited stress can promote psychological growth to some extent, high levels of and/or prolonged work-related stress can adversely affect both mental and physical health [4,5]. When it persists over an extended period, work-related stress can lead to exhaustion, burnout, cynicism, and apathy, making it difficult for the affected individual to cope with challenges [6]; it is one of the strongest factors influencing job satisfaction and burnout symptoms [7].

The ICD-11 defines burnout as a syndrome resulting from chronic workplace stress that has not been successfully managed. It is characterized by three dimensions: feelings of energy depletion or exhaustion, increased mental distance from or negative feelings associated to one's job, and reduced professional efficacy.

The reported incidence of burnout ranges largely due to the heterogeneity of measurement methods and because it is not classified as a medical condition and therefore not systematically monitored [8,9]. Even within populations with the same profession, such as physicians, we find prevalences ranging from 0% to 80.5% [10]. A meta-analysis from Switzerland reported a 4% prevalence of severe burnout and an average burnout prevalence of 18% [8]. Regarding professions that were under particular pressure due to the COVID-19 pandemic, such as healthcare professionals, the aggregated proportion of reported burnout was 39% [11]. In the USA in 2016, expenditures for the consequences of work-related stress ranged from $125 to $190 billion dollars a year [12].

In recent decades, there has been continuous innovative development in the field of information technology, with these milestones contributing to the advancement of e-healthcare in terms of quality, continuity, and efficiency [13]. Natural Language Processing (NLP), a field that has evolved to become mostly a subset of the broader field of Artificial Intelligence (AI), has made rapid progress in the past two decades. NLP tools are now used daily by every smartphone user and have significantly contributed to the development of speech translation, chatbots and personal digital assistants, and voice-controlled home automation systems [14]. NLP can also be used for text mining tasks, drawing useful insights from unstructured text data [15]. In particular, text mining also has different applications and use cases in the health domain, see e.g., [16]. Various techniques and algorithms are used in text mining, including supervised and unsupervised methods for document categorization and cluster analysis [17].

Recent studies have shown promising results in the use of NLP for the detection of burnout syndrome. One study achieved an accuracy of close to 80% in detecting burnout from text snippets using support vector machine (SVM) classifiers [18], providing a foundation for future generations of clinical methods based on NLP [18]. A later study analyzed the performance of NLP methods using a larger dataset and improved upon the first study by achieving a balanced accuracy of 0.93 and a test recall of 0.93 [19]. Another study also demonstrated the potential of machine learning methods in predicting early indicators of job burnout in employees, achieving a 70% prediction accuracy [20]. A systematic review of literature from 2015 to 2020 demonstrates that information technologies, particularly software and mobile apps, lead to improved detection of burnout syndrome [21]. The authors suggest considering the use machine learning for predicting susceptibility to burnout, aiding in preventive measures for companies. The review recommends developing mobile apps incorporating the Maslach Burnout Inventory test for accurate symptom assessment and suggests further research on technology's role in burnout-related conditions [21]. These studies collectively highlight the potential of technology and in particular NLP and machine learning in the early detection of burnout syndrome.

Natural Language Processing based on textual data has also been applied to detect indicators for other mental health problems from textual data. A recent survey [22] gives an overview of the work in the field, and shows that there are several limitations in the state-of-the-art:
a) A vast majority of the data used to train the classifiers are based on data from social media. Even though good performance metrics are achieved, this can lead to major limitations when applying these systems to real-world patient data, including (1) statements about diseases being subjective, (2) little or no demographic information and hence difficulty in ensuring diversity in the datasets, or (3) restrictions on the sharing of collected social media data for research [22].
b) A large part of research (45% of the examined studies in [22]) focused on depression, demonstrating an opening for researchers to explore other issues, such as burnout or stress.
c) The vast majority (81%) of the papers surveyed in [22] dealt with English texts, with 10% and 1.5% including Chinese and Arabic respectively. As language is very relevant in the context of mental health [22], addressing additional languages in use of NLP for mental health is essential. In general, the strong bias towards relatively few languages in NLP research is criticized in the literature [23].

Explainability is crucial in NLP applications for mental health, in order to ensure the robustness, trustworthiness and effectiveness of the models. Explainability refers to the ability to describe the internal processes of a model in a way that is understandable to humans. Given the sensitive nature of mental health data, it is essential for stakeholders, including healthcare professionals and patients, to understand how and why certain decisions or predictions are made by the models. Such transparency can help in validating the model's reliability, identifying potential biases, and making informed adjustments to improve accuracy and fairness.

In the field of mental health, Wang [24] investigated interpretable and explainable AI models for Alzheimer's, Dementia, and Depression. The study introduced an interpretable deep learning framework designed to automatically detect Alzheimer's and Dementia using patient interview transcripts. This framework used attention mechanisms to emphasize the sections of the transcripts that had the most significant impact on the model's decision-making process. By concentrating on specific linguistic features and their relevance to the diagnosis, the model offered insights into the correlation between certain speech patterns, Alzheimer's and Dementia.

Han et al. [25] introduced a Hierarchical Attention Network for explainable depression detection on Twitter. Their model uses metaphor concept mappings and incorporates a novel attention-based encoder to enhance interpretability. By emphasizing both context-level and word-level features, this approach offers valuable insights into how specific tweets and metaphors are employed by individuals with depression, thus providing additional justification for the model's predictions.

Large language models (LLMs) have shown great promise in various applications due to their ability to process and generate human language. Wang et al. [26] demonstrated the efficacy of LLMs in not only predicting but also explaining responses to the Beck's Depression Inventory (BDI) questionnaire using social media data. The study highlighted the ability of LLMs to provide transparent and interpretable results, which are critical for clinical applications.

The main objective of this study is to compare the performance of BERT text classifiers in detecting indicators for burnout fine-tuned on different data sources, including a) online data, commonly used in the state-of-the-art, b) a manually curated synthetic dataset, and c) an anonymously collected real-world dataset. In addition, we conducted a qualitative analysis of explainable AI methods applied to the trained models in an interdisciplinary focus group.

## Methods

We applied different methods for data triangulation of (1) pre-existing online datasets, (2) manually curated synthetic datasets, and (3) a real-world dataset featuring free-text responses alongside a validated burnout scale. The labels predicted by the machine learning models on the real-world data and features highlighted by explainable AI methods were discussed in consultation with an interdisciplinary focus group consisting of domain experts.

### Dataset 1: Online Data (Baseline)

As a baseline, we use an existing dataset [27] consisting of a mixture of interviews with burnout patients from Rahner [28] and web scraped data. Given that descriptions of burnout experiences in German are scarce (as opposed to English, see e.g., Merhbene [19] using Reddit data), manual scraping and data curation was used. Various data sources were considered in data collection: online articles, blogs, interviews containing testimonials from people affected by burnout, and transcripts from Youtube videos in which burnout experiences were recounted. These online sources were used both for the burnout data, as well as the control group (testimonials of people describing their daily life with regard to their job, without the matter of burnout being relevant). Further statements from the book Buchenau [29], which contains 13 testimonials of people who have suffered from burnout syndrome, were added. As some texts of this book seemed like to have undergone proof-reading and redaction, examples not too heavily stylized were selected.

The resulting dataset contained 310 records for the control class and 387 for the burnout class, which were manually curated. We use this dataset as a baseline in this paper, as an example indicative of the vast body of research using online data in the field of NLP for mental health.

### Dataset 2: BurnoutExpressions

Given the limitations of using web-scraped data for real world applications, we decided to explore the potential of state-of-the-art text generation models, such as the latest GPT models, to create the BurnoutExpressions dataset. The following procedure was used to create this dataset: Initial seed words or expressions (N=103) describing the symptoms of burnout were extracted from the symptoms overview table presented by Albrecht [30], which was built upon the work of Burisch [31]. In some cases, the burnout symptoms described in the table are contradictory (e.g., hyperactivity and lack of energy). Emphasis was placed on expressions that seem likely to be expressed in text form by patients answering open questions relating to their personal experience. Some symptoms, such as rigid black-and-white thinking (de: "rigides Schwarzweissdenken"), are more difficult to capture in only a few words or phrases.

The table of symptoms covers aspects ranging between motivational issues like *reduced engagement* (de: reduziertes Engagement), emotional topics such as *loss of positive feelings* (de: Verlust positiver Gefühle), and physical symptoms like *back pain* (de: Rückenschmerzen).

As detailed in the following section, a native German speaker augmented the expressions extracted from the symptom table with synonyms and additional context. The data for the control group was created by selecting an opposing word or phrase for each expression in the burnout list (e.g., simply inserting the word *not* to negate expressions indicating symptoms of *burnout).* Together, these expressions make up the BurnoutExpressions v1 dataset (N=508 for burnout, and N=508 for the control group).

This manually curated dataset was then augmented using generative AI (OpenAI API with GPT-3.5 Turbo). Sentences were generated based on the expressions from the initial dataset, leading to the BurnoutExpressions v2 dataset (N=2026 for burnout, and N=1895 for the control group), which is descrbed in further detail below. Figure 1 provides an overview of the dataset compilation process.

The work presented here relies on binary classification for burnout vs. no burnout, in line with the majority of work in the field of NLP for the detection of mental health issues (e.g., Tadesse et al., 2019 [32] for depression; Shen and Rudzicz, 2017 [33], for anxiety) as well as commonly used inventories, which typically make use of a threshold score for classification. In a later section, we discuss the potential of more granular classification based, for example, on the dimensions underlying clinically validated inventories, which arose as a salient topic within a focus group of medical experts and the limitations of this study.

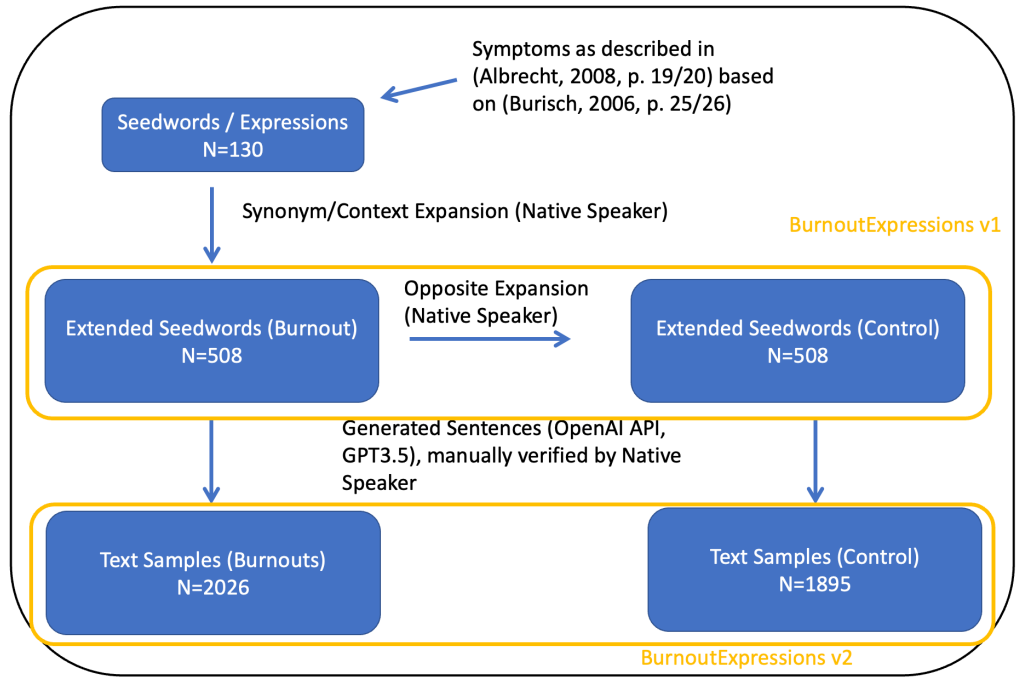

*Figure 1 The overall process to create the BurnoutExpressions dataset.*

### Dataset 2a: BurnoutExpressions v1 (Manually Curated Dataset)

For each word or expression from the initial list (based on the table from Albrecht [30]/Burisch [31]), synonyms or words with a similar meaning (expanding the context) were manually collected by a native speaker. A focus was placed on formulating self-narratives based on selected terms (e.g., *ich bin heute nicht da* (engl: *I am not here today*) in relation to the word *Fehlzeiten (*engl: absences)), or rephrasing expressions in a more colloquial form using infinitives (e.g., *keine Hoffnung mehr haben* (engl: *not to have hope anymore)* as a variant of *Desillusionierung* (engl: *Disillusionment*)). Some phrases expressing actions based on seed words were used; for example, from *Neigung zum Weinen (*engl*: Tendency to cry)*, the expressions *nahe am Wasser gebaut sein[1]* or *Tränen fliessen* (engl: *tears flowing*) were derived. For some expressions, several additional variants were identified, in other cases, no new phrases were added. All terms were aggregated and checked for duplicates. This resulted in an extended list of 508 unique expressions concerning the symptoms of burnout.

---

[1] Word by word translation to English: *being built near the water.* Meaning of the expression in English: *Being a person that cries easily.*

To obtain a list of expressions for data representing the lack of burnout indicators, for each element of the 508 burnout expressions, an opposing phrase was manually identified by a native German speaker. The identification of opposites proved challenging in some cases; some words did not have a clear opposite, or, as mentioned above, some symptoms from the literature were already oppositional. Not only exact opposites were considered; more general terms describing the absence of a particular symptoms were also used. For example, it was difficult to find an opposite for *Magen-Darm-Geschwüre (*engl*: Gastrointestinal ulcers)*, so a more general expression like *sich gesund fühlen* (engl: *feeling healthy*) was selected.

The combined BurnoutExpressions v1 dataset consists of 508 expressions with the label 0 for the control group and 508 with the label 1 for the burnout group. Table 1 gives examples from both categories.

*Table 1 Examples from BurnoutExpressions v1 dataset.*

| Burnout Expression | Control Group Expression |
| --- | --- |
| emotionale Erschöpfung (engl: emotional exhaustion) | emotionale Kraft haben (engl: have emotional power) |
| Abstumpfung (engl: Blunting) | Euphorie (engl: Euphoria) |
| Stimmungsschwankungen (engl: Mood fluctuations) | ausgeglichen sein (engl: be balanced) |
| Nörgeleien (engl. Nagging) | positives Feedback geben (engl: give positive feedback) |
| … | … |

Dataset 2b: BurnoutExpressions v2 (Manually Curated + GPT Text Generation)
In a second step, the BurnoutExpressions v1 dataset was extended using generative AI. Additional sentences were generated using OpenAI's API with the GPT3.5 model (gpt-3.5-turbo-1106). The initial dataset was processed in batches of 20 entries (all burnout and all control group expressions were iterated). The prompt was constructed as follows:

messages=[{"role": "user", "content": prompt}
with
prompt = "Generate 10 sentences each in German for the following expressions. The sentences should represent the wording of a person being in this kind of mental state:» + the 20 words from a specific batch separated by comma.

In some cases, sentences were partially cut (in cases where longer sentences were created), or fewer sentences were generated than requested by the prompt. Any partial sentences were removed manually.

After this step, the BurnoutExpressions v2 dataset was compiled, containing N=2026 sentences with burnout symptoms, and N=1895 sentences without burnout symptoms. Table 2 shows an example of the generated data.

*Table 2 Examples from the BurnoutExpressions v2 dataset.*

| Text | Label |
|---|---|
| Ich spüre eine große emotionale Kraft in mir, die mich antreibt. *(engl: I feel a great emotional strength within me that drives me forward.)* | 0 |
| Trotz der schwierigen Situation bewahre ich meine emotionale Kraft. *(engl: Despite the difficult situation, I retain my emotional strength.)* | 0 |
| Ich fühle mich voller emotionaler Energie und Tatendrang. *(engl: I feel full of emotional energy and drive.)* | 0 |
| Der Stress hat mich längerfristig in einen Zustand der Erschöpfung versetzt. *(engl: The stress has put me in a long-term state of exhaustion.)* | 1 |
| Meine körperliche Erschöpfung ist so groß, ich fühle mich wie eine leere Hülle. *(engl: My physical exhaustion is so bad, I feel like an empty shell.)* | 1 |
| Die geistige Erschöpfung macht es mir schwer, klar zu denken. *(engl: The mental exhaustion makes it difficult for me to think clearly.)* | 1 |
| … | … |

### Dataset 3: Real-World Dataset for the Test Phase

A real-world dataset was compiled for validation by conducting an anonymous online survey and to discuss the labels assigned by the machine learning models with clinical and scientific experts in the focus groups described later.

*Recruitment online survey*

A convenience sampling among adults from the German-speaking part of Switzerland was conducted. Emails with information about the study's aims, inclusion criteria, data protection, and the survey link were dispersed amongst the researchers' networks, along with a request to forward the invitation to other colleagues. The researchers also posted the study information on social media platforms (LinkedIn and X) and on mental health-specific community platforms such as www.psychic.de. Participation was voluntary.

### Data collection

Cross-sectional data was collected using the online survey instrument Limesurvey between August and November 2023. The instrument asked participants to provide their age before proceeding to answer four free-response questions developed by the researchers, as well as the 16-item German Version of the Oldenburg Burnout Inventory (OLBI) [34]. Thus, each respondent provided four text samples labelled with a clinically validated burnout score.

### Free-text fields

The four free-text fields were developed to allow participants to describe how they feel on a usual day. The free-text questions were pretested using cognitive debriefing with five individuals from the researchers' networks. They found the questions to be understandable and answerable as intended. The free-text questions are detailed in Appendix 1. In a few cases, test subjects provided no or single word responses; these cases were removed from the dataset.

### Oldenburg Burnout Inventory

The Oldenburg Burnout Inventory measures burnout along two dimensions, namely "disengagement from work" (disengagement) and "exhaustion" (exhaustion). Clinical studies have shown the inventory to be valid and reliable [35]. The items were scored on a 4-point Likert scale ranging from "Strongly agree" (1) to "Strongly disagree" (4). The aggregate score for the inventory is the mean of the respondent's answers.

Different cut-off values are suggested in the literature to translate the scores along the different dimensions of the OLBI into a burnout or burnout risk indicator. In this study, we considered three different thresholds:

The first threshold classifies burnout as corresponding to scores ≥2.25 for exhaustion and ≥2.1 for disengagement [36]. Researchers have demonstrated that these values correlate to the mean scores on the widely used Maslach Burnout Inventory (MBI) amongst employees diagnosed with burnout by a physician [37].

The second threshold differentiates between very high burnout in a working sample and high burnout in a clinical sample. For the working sample, the cut-off scores are ≥2.85 for exhaustion and ≥2.6 for disengagement. For the clinical sample, the threshold scores are higher, with cut-offs of ≥3.13 for exhaustion and ≥2.72 for disengagement, reflecting more severe burnout [38].

The third threshold in the literature uses total scores in place of means, with a score of 35 or above (out of 64) indicating an increased risk for burnout. This approach has been referenced in the literature as a reliable indicator of heightened burnout risk [39].

### Focus Group

A focus group was organized with clinical and scientific experts in mental health to validate the labels predicted by our machine learning models on samples from the real-world dataset. Participants were recruited by convenience sampling from the researcher's networks.

We planned 1 ½ hours for the focus group, in which the results from dataset 3 were presented and discussed with regard to plausibility and implications.

In the focus group, we presented samples of model predictions and asked the participants to rate whether they agree with the AI's output. For each sample the proportion of agreement was calculated. In addition, the reasons for rejection were collected and discussed. We also included open questions to initiate a discussion among the participants on their experiences of how burnout symptoms are expressed verbally. The focus group interview was documented in writing by an assistant and the protocol was analyzed using content analysis.

### Data Analysis

We used Dataset 3 as validation data for machine learning classifiers trained on the (1) the online dataset, (2) BurnoutExpressions v1, (3) BurnoutExpressions v2, and (4) a combined dataset of the online dataset and BurnoutExpressions v2. We wanted to investigate how classifiers trained on non-clinical data perform on real-world data.

### Experimental Setup

The overall experimental setup consists of the following steps: a) selecting a pre-trained BERT model for the German language, b) adding vocabulary not priorly known to the model (e.g., words from our BurnoutExpressions v1 dataset), c) fine-tuning the resulting model for binary classification (indication for burnout in the text, or not) using a random split with 80% of each dataset for training (this step is done separately for the four datasets, resulting in four classifiers), d) testing the classifiers with real-world data obtained from a survey (text answers labelled with the Oldenburg Burnout Inventory scores). The experimental setup is visualized in Figure 2.

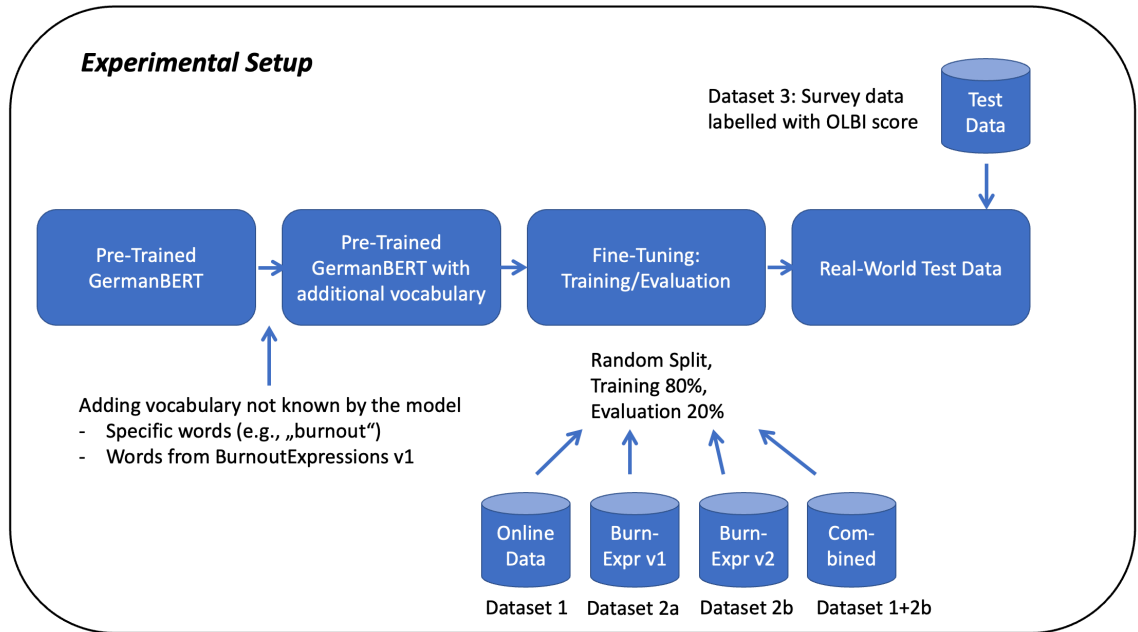

*Figure 2 Experimental Setup: the different steps of training 4 burnout classifiers, and testing them with real-world data.*

### Training/Evaluation: GermanBERT Classifiers

As a foundation model for the four classifiers, we use the GermanBERT model version *bert-base-german-cased*[2]. Any words from the BurnoutExperessions v1 that were not already in the model's vocabulary were added before fine-tuning. We then used *BertForSequenceClassification* to fine-tune the model for binary classification using the HuggingFace library[3]. The training arguments are detailed in Appendix 2. We experimented with 2-4 training epochs, all but one of the final models were trained for 3 epochs, with the remaining model taking only 2 epochs.

Using this setup, we trained four models with different training data: (1) for the online data, (2) for BurnoutExpressions v1, (3) for BurnoutExpressions v2, and (4) a combined dataset of (1)+(3).

### Test Datasets: Applying Classifiers to Real-World Data

Dataset 3, collected with a survey and taken to be a good representation of real-world data, was used to test the fine-tuned models, but never as training data. Each test subject's answers to each of the free-response questions was labelled by the model in question according to whether it contained indicators for burnout.

Three different labels based on the effective OLBI scores, one for each of the cut-off values specified in the literature (see section Method for details), were considered as the ground truths and compared to the classifiers' predictions.

---

[2] https://huggingface.co/google-bert/bert-base-german-cased
[3] https://huggingface.co/transformers/v3.0.2/model_doc/bert.html#bertforsequenceclassification

*Explainability Study*

To get an indication of what information the system used in its decision-making process, an explainability study using the library transformers-interpret [4] was conducted. The predictions of the classifier trained on the combined dataset applied to the real-world data from the survey were examined. Using transformers-interpret, visualizations were generated of the weight the model attributed to each word in a given sample, indicating the contribution of given words towards or against the classifier's prediction (indication for burnout, or not). These visualizations were then discussed with domain experts in focus groups.

*Ethical Considerations*

An ethics approval is deemed unnecessary according to the legislation Swiss Federal Act on Research Involving Human Beings (see 810.30 HRA Art. 2) since data collection was anonymous for the online data and in the focus group interview no health related data of the participants was collected. The study was conducted in accordance with the Declaration of Helsinki. It was performed on a voluntary basis for all participants. All participants were free to stop filling out the questionnaire at any time. Participants received written information before the start of the study about the contents, aim and voluntary nature of their participation and gave their informed consent by completing the first survey page, respectively, by participating in the focus group interview.

## Results

### Training: GermanBERT as Classifier

For each classifier, the training data was split into 80% for training and 20% for evaluation (random split). The following graphs show the training loss, as well as the loss, f1-score and accuracy from evaluation. We trained four classifiers for the four different datasets as mentioned in the experimental setup (see Figure 2).

In Figure 3, we observe the performance of the GermanBERT classifier which was trained on data collected from online sources. The training loss (red line) consistently decreases, indicating effective learning. The evaluation loss (blue line) also decreases, closely mirroring the training loss, which suggests good generalization to the evaluation data. The F1-score (orange line) and accuracy (green line) both show steady improvement, stabilizing around 0.85 and 0.8, respectively. These trends demonstrate that the classifier is effectively learning to distinguish between burnout and non-burnout cases, achieving high accuracy and F1 scores with low loss values by the end of the training period.

---

[4] https://github.com/cdpierse/transformers-interpret

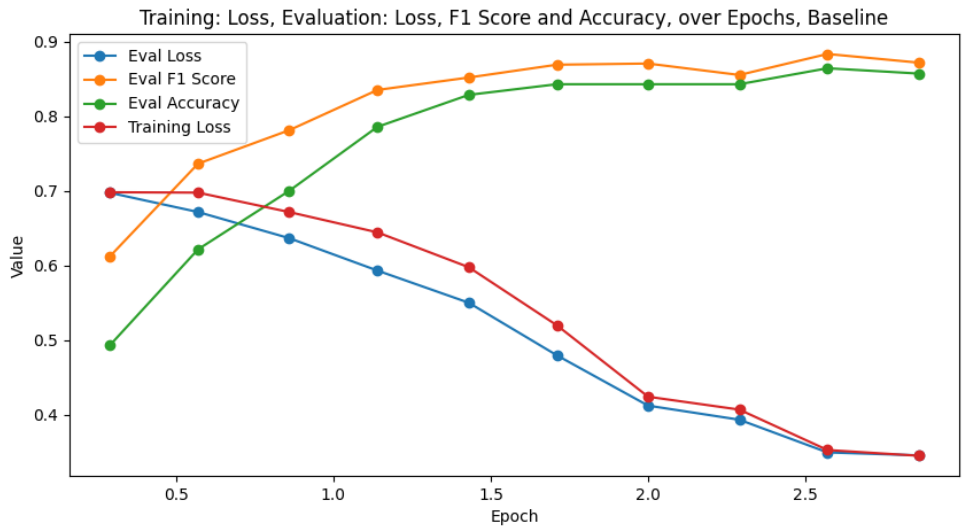

*Figure 3 As a baseline, we consider the classifier trained on the data collected from online sources.*

In Figure 4, we observe the performance of our classifier trained on the BurnoutExpressions v1 dataset. The training loss (red line) shows slight fluctuations but generally trends downward, indicating that the model is learning. The evaluation loss (blue line) remains relatively stable, with a slight decreasing trend, suggesting consistent performance on the evaluation data without significant overfitting. The F1-score (orange line) starts low and rapidly increases, stabilizing around 0.65, which indicates an improved balance between precision and recall. Similarly, the evaluation accuracy (green line) shows a steady increase initially, stabilizing around 0.65, reflecting the model's growing ability to correctly classify burnout cases. However, the performance of the GermanBERT model trained on online data still outperforms that of the model trained on the BurnoutExpressions v1 dataset.

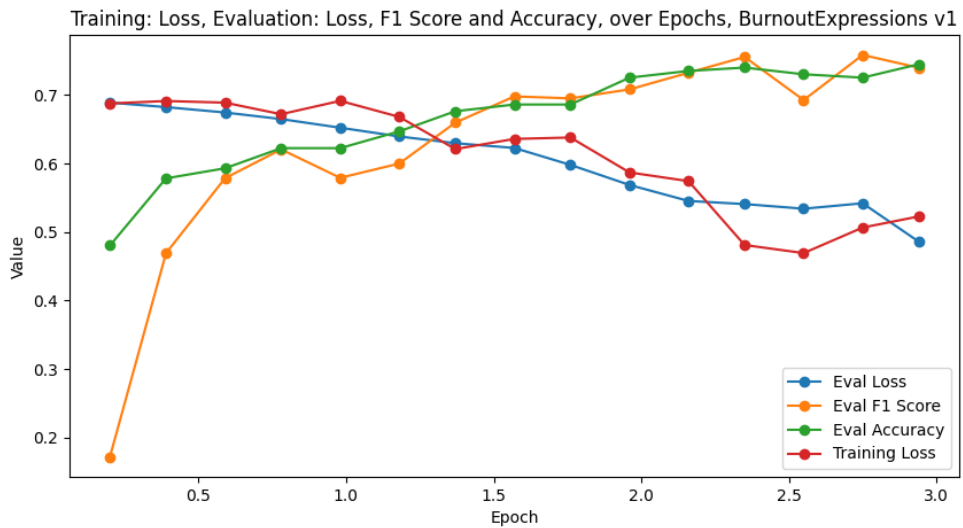

*Figure 4 Training and evaluation performance of the classifier trained on the BurnoutExpressions v1 dataset.*

The training and evaluation performance of the GermanBERT classifier trained on the BurnoutExpressions v2 dataset is illustrated in Figure 5. The training loss (red line) shows a consistent decline, indicating the model's effective learning over time. The evaluation loss (blue line) also decreases, mirroring the trend of the training loss, which suggests that the model generalizes well to the evaluation data without significant overfitting. The F1-score (orange line) and accuracy (green line) show substantial improvements, quickly rising and then stabilizing around 0.85 and 0.83, respectively. This trend demonstrates that the classifier effectively learns to distinguish between burnout and non-burnout cases, achieving high accuracy and F1 scores with low loss values by the end of the training period.

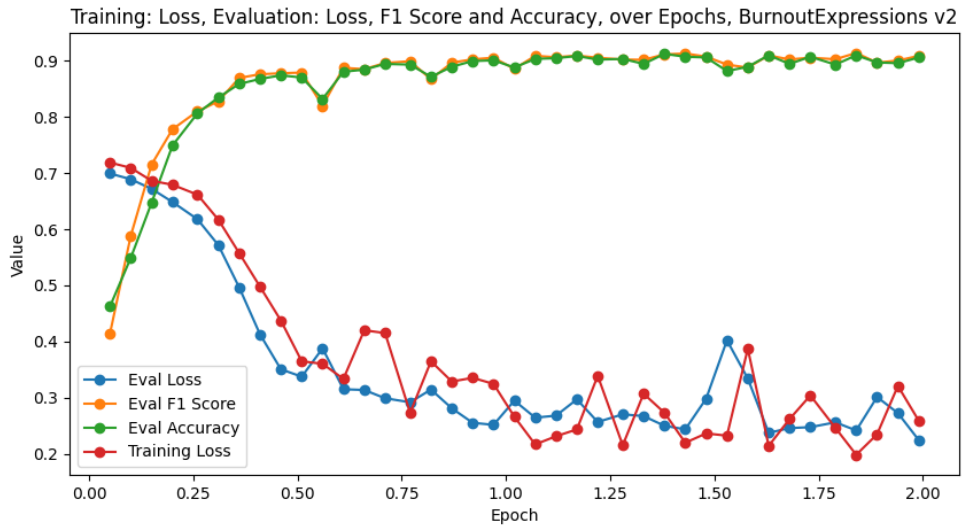

*Figure 5 Training and evaluation performance of the classifier trained on the BurnoutExpressions v2 dataset.*

Figure 6 presents the training and evaluation performance of the GermanBERT classifier trained on the combined dataset, which includes both online data and BurnoutExpressions v2. The training loss (red line) demonstrates a continuous downward trend, signifying the model's effective learning process. The evaluation loss (blue line) follows a similar downward trajectory, further confirming the model's ability to generalize well to unseen data. Both the F1-score (orange line) and accuracy (green line) show rapid increases initially, stabilizing at approximately 0.87 and 0.85, respectively. These metrics indicate that the classifier trained on the combined dataset achieves similar performance compared to the model trained solely on the BurnoutExpressions v2 dataset.

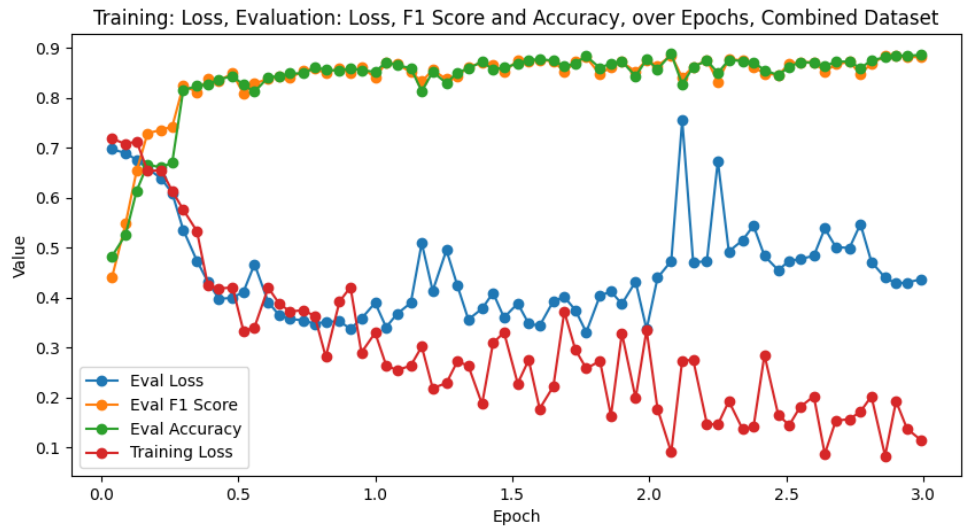

*Figure 6 Training and evaluation performance of the classifier trained on the combined dataset (online data + BurnoutExpressions v2).*

### Test Phase: Real-World Dataset

The real-world dataset is composed of the answers to free-text questions and answers to the OLBI questionnaire. Each free-text answer was labelled according to the three different cut-off values for OLBI (see method for detailed description). In total, N=17 individuals (8 female, 8 male, 1 no gender specified) completely filled out our online survey. Each text provided (i.e., the answers to each free-text question, considered separately) was mapped to the OLBI score. Empty answers were dropped (2 cases). This resulted in 66 data points. Based on the OLBI scores of each individual, the labels 1 (indication for burnout) and 0 (no indication for burnout) were attributed to each data point according to three different threshold values. This led to the following distribution shown in Table 3.

*Table 3 Distribution of Burnout and No Burnout Labels based on the different cut-off values.*

| Cut-Off Value | Nr. Burnout (Label 1) | Nr. No Burnout (Label 0) |
|---|---|---|
| Cut-Off Value 1 | 4 | 13 |
| Cut-Off Value 2 | 2 | 15 |
| Cut-Off Value 3 | 7 | 10 |

Even though the survey was intensively distributed among people potentially suffering from a burnout or related mental health problem, only a small subset of respondents had an OLBI score high enough to be classified as part of the burnout group according to each of the cut-off scores.

We then applied our four machine learning classifiers to these samples. Table 4 gives an overview of the F1-Scores we observed in our experiments.

Table 4 Overview of the F1-Scores from the different experiments on Dataset 3.

| Pre-Trained Model | Dataset for Fine-Tuning | Epochs | F1 for Cut-Off 1 of Test Data | F1 for Cut-Off 2 of Test Data | F1 for Cut-Off 3 of Test Data |
|---|---|---|---|---|---|
| GermanBERT | Online Data (Baseline) | 3 | 0.170 | 0.130 | 0.230 |
| GermanBERT | BurnoutExpressions v1 | 3 | 0.360 | 0.570 | 0.290 |
| GermanBERT | BurnoutExpressions v2 | 2 | 0.330 | 0.350 | 0.440 |
| GermanBERT | Combined Dataset: BurnoutExpressions v2 + Online Data | 3 | **0.580** | **0.610** | **0.510** |

### Explainability Study and Focus Groups

Table 5 summarises the examples provided, the model's predictions and the proportion of experts in agreement. The first example was classified as below the threshold for burnout. All experts agreed and stated that, in particular, the phrase "recharged energy" was indicative of not having burnout. However, it was discussed that the expression could relate to only a snapshot of a specific moment and should be considered in the context of a more extended period. The sentence "longing for more free time" may indicate a burnout risk factor and was also marked as a indicating potential risk by the AI model. However, the positive statements in the sample in question were rated more relevant. For the second example, most agreed with the labelling that the sentence indicates burnout. However, it was discussed whether such statements may not be somewhat indicative of depression rather than burnout. The line between the two is not clear-cut, leading to a challenge identification. For the third example, the experts agreed it should not be labelled as burnout because of missing information, which is crucial for a clear allocation. It is unclear which aspects of the day resulted in the reported exhaustion; burnout is a relates to the occupational context, but work-related information was not provided. This is also the case for example four, which was labelled as burnout by AI, whereas the participant did not reach any OLBI threshold for burnout. In this example as well, no work-related information is given.

Table 5 Examples (translation from German) with OLBI cut-off score, AI-assigned label and expert agreement.

| ID | Example | OLBI Cut-off | Labelled by AI | Expert Agreement |
|---|---|---|---|---|
| 1 | It was a very relaxing weekend that recharged my energy. Cheerful mood had an impact on the experience. More relaxed and spontaneous as there were | No burnout | No burnout | 100% |
| | no obligations and time pressure. Longing for more free time | | | |
| 2 | I go to bed. Because I'm completely exhausted. I don't care about the day. It's like the next one or the one before. A struggle from morning to night | burnout | burnout | 80% |
| 3 | I am often exhausted from the day and feel more or less satisfied with my daily activities. Most of the time, I plan things that I want to do better in the future. I try to reflect on which strategies I can use to create more meaning and satisfaction in my life | 2/3 burnout | No burnout | 100% |
| 4 | My partner was away over the weekend, and I felt partly lonely, longing for a 'real' family. I was able to positively influence my mood by doing something with the little ones and meeting another family. Through what I did, I could influence my feelings and mood positively. Conversely, my emotions and mood led to me perhaps needing a bit more energy to get things done. | No burnout | burnout | 0% |

## Discussion

### Principal Results

On one hand, the classifier trained on web data shows poor performance in our experiments, despite demonstrating high performance in the training and validation phases (see Figure 3). On the other hand, considering that the real-world test dataset is imbalanced (see Table 3), the performance can be seen as moderate for the classifiers presented in this paper. In particular, the best f1-scores were observed using the combined dataset, which was training on online data, manually curated data and AI-generated synthetic data. This indicates that the wider scope of the combined data allows the classifier to better handle unseen data.

The manually curated dataset based on the symptoms described in the literature and then extended with generative AI performs well during training and gives generally promising results on validation data. Interestingly, there is not much difference in performance between the manually curated dataset, and the synthetically augmented data extending it (BurnoutExpressions v1 even performs better in two cases, see Table 4). We speculate that the generated texts provide greater volume, but not more information. The results from the focus group indicate that there is high agreement with the decision of the labelling. However, the experts expressed several limitations, which should be considered to increase the model's performance and reliability.

### Comparison with Prior Work

We observed high validation F1-scores in the training phase for all datasets, up to over 0.9. This is in line with existing work in the field. For example, similar performance was reported for burnout detection on Reddit data [19], or for Depression detection [32]. However, our results on the real-world test dataset indicate this performance does not necessarily transfer to other, closely related types of data. Whereas the other work in the field of mental health classification uses social media data in the vast majority of cases (81% in a recent survey paper [22]), these models might not be readily transferable to clinical data. Other recent work in the field has developed specific language models specialized for the mental health domain, including MentalBERT [40] and MentaLLaMA [41]. However, these are primarily trained on English data and thus cannot be compared directly to our results, dealing with German text data.

### Burnout vs. Depression

Existing literature shows that the empirical evidence for distinguishing burnout from depression is inconsistent [42]. A meta-analysis indicated a high association between burnout and depression, as well as anxiety [43]. This overlap of symptoms makes it a challenge to make a clear-cut diagnosis. Moreover, the evidence of discriminant validity between burnout and depression is low, theoretically and empirically. One clear distinguishing criterion is that burnout is currently understood as a phenomenon in the occupational context [44], which may be the key discriminating factor. When applying machine learning, the underlying measurement instrument, seen as a valid criterion, must be chosen critically. However, the discourse has been further complicated by the emergence of parental burnout as a burnout phenomenon that is not work-related [45]. There seem to be some specific consequences of the conditions which allow differentiation, such as the intent to leave the organization for job burnout or violence/neglect for parental burnout, none of which are fully explained by depression in general [45]. A text-based assessment using AI should therefore take such behaviors into account when collecting data.

### Snapshot vs. Long-Term

This study used texts that are situational excerpts from people's everyday lives. A person who writes that they are exhausted may feel exhausted, but it does not necessarily have to be directly associated with burnout. Research shows that stress reactions do not lead directly to burnout, but rather, burnout is a response to chronic occupational stress [42]. It concerns the amount of time certain stress factors are experienced at work and resist resolution. The time factor makes capturing data using text even more challenging because retrospection is often distorted. Research indicates that memory bias is subject to negative mood states. In particular, feelings of anxiety, depression, and helplessness are amplified in overall retrospective assessments compared to daily ratings [46]. The text should therefore not be collected at a single point in time but several times over a certain period from the same person to create a more adequate data quality.

### Limitations

One limitation of our work is the small sample size of the real-world dataset. Additionally, the dataset was unbalanced. Our findings will need to be confirmed with a larger dataset. Our survey was conducted anonymously, and thus we cannot assess how well the classification has worked for different demographic groups; this needs to be further explored to avoid encoding societal bias within machine learning models. Modeling mental health as a binary classification task is oversimplified, and collaborators in clinical psychology have expressed a desire for future work to explore the potential of more fine-grained classification, taking into assessing different dimensions of burnout described in the literature, considering specific symptoms, and integrating comorbidity and comparison with other mental health issues. Reproducing these kinds of experiments is challenging, as the fine-tuning phase of BERT models is a stochastic process, and the randomized test sets were small enough that performance metrics exhibited significant variance. In future work, it could also be interesting to incorporate methods that improve model performance by automatically isolating sentences in the data that are the most relevant for assessing mental health, similar to approaches in earlier work [47] in the context of eating disorders. Such targeted scraping methods could also be used to augment our existing dataset with similar sentences from online discussion forums and other relevant corpora.

### Conclusions

Our research indicates that further validation of the state-of-the-art with real-world data is required to ensure the methods based on online data transfer to real clinical data. This is enabled by an interdisciplinary collaboration, where NLP and healthcare researchers work in close collaboration with clinical partners.

### Acknowledgements

This research was supported by funding from the thematic field Caring Society from the Bern University of Applied Sciences, which we gratefully acknowledge. We would like to express our gratitude to Sophie Haug for providing the dataset used in this study, which was produced as part of her bachelor thesis at Bern University of Applied Sciences.

### Conflicts of Interest

None declared.

### Abbreviations

JMIR: Journal of Medical Internet Research
ML: Machine Learning
NLP: Natural Language Processing
OLBI: Oldenburg Burnout Inventory

RCT: randomized controlled trial

## Appendix 1 Survey Free-Text Questions

**Question 1**
Beschreiben Sie, wie Sie sich an einem gewöhnlichen Wochentag fühlen, wenn Sie morgens aufstehen. Beschreiben Sie Ihre Gefühle oder die Art und Weise, wie Sie Ihren Körper wahrnehmen. Versuchen Sie, schriftlich auszudrücken, wie Sie den Tag angehen.
*(engl: Describe how you feel on a normal weekday when you get up in the morning. Describe your feelings or the way you perceive your body. Try to express in writing how you approach the day.)*

**Question 2**
Wie fühlen Sie sich während des Mittagessens an einem typischen Wochentag? Beschreiben Sie, wie Sie diese Tagesmitte erleben und welche Gefühle Sie in Bezug auf Ihren Alltag haben.
*(engl: How do you feel during lunch on a typical weekday? Describe how you experience this middle of the day and what feelings you have in relation to your everyday life.)*

**Question 3**
Beschreiben Sie Ihre Gefühle vor dem Schlafengehen an einem typischen Wochentag. Wie blicken Sie auf den Tag zurück, den Sie erlebt haben? Beschreiben Sie Ihre positiven und/oder negativen Eindrücke. Wie haben Sie sich gefühlt, haben sich diese Gefühle im Laufe des Tages verändert? Wie haben Sie sich und Ihre Umgebung wahrgenommen?
*(engl: Describe your feelings before going to bed on a typical weekday. How do you look back on the day you experienced? Describe your positive and/or negative impressions. How did you feel, did these feelings change during the day? How did you perceive yourself and your surroundings?)*

**Question 4**
Denken Sie an das letzte Wochenende. Wie haben Sie sich gefühlt? Beschreiben Sie, wie Sie das Wochenende erlebt haben und in welcher Stimmung Sie es wahrgenommen haben. Haben Ihre Gefühle oder Ihre Stimmung die Art und Weise beeinflusst, wie Sie die Dinge erledigt haben?
*(engl: Think back to last weekend. How did you feel? Describe how you experienced the weekend and the mood in which you perceived it. Did your feelings or mood influence the way you did things?)*

## Appendix 2 Training Parameters

The following parameters were used for the training of the four classifiers:

```
training_args = TrainingArguments(
    num_train_epochs=3,
    per_device_train_batch_size=16,
    per_device_eval_batch_size=64,
    warmup_steps=500,
    weight_decay=0.01,
)
```

With exception for the dataset BurnoutExpressionsv2, where num_train_epochs=2 was used instead.